\documentclass[letterpaper, 10 pt, conference]{ieeeconf}  % Comment this line out if you need a4paper

\IEEEoverridecommandlockouts                              % This command is only needed if 
\usepackage{graphicx} % for pdf, bitmapped graphics files
\usepackage{amsmath} % for \boldsymbol macro
\usepackage{cases}%

\title{\LARGE \bf
An Electrocommunication System Using FSK Modulation and Deep Learning Based Demodulation for Underwater Robots
}

\author{Qinghao Wang$^{1}$, Ruijun Liu$^{1,2} $, Wei Wang$^{3,4,\ast}$, and Guangming Xie$^{1,5,6,\ast}$
\thanks{$^{1}$State Key Laboratory of Turbulence and Complex Systems, Intelligent Biomimetic Design Lab, College of Engineering, Peking University, Beijing, 100871, China.
  }%
\thanks{$^{2}$College of Electrical and Information Engineering, Guangxi University of Science and Technology, Liuzhou, 545006, China.
        }%
\thanks{$^{3}$Department of Urban Studies and Planning, Massachusetts Institute of Technology, Cambridge, MA 02139, USA.
}
\thanks{$^{4}$Computer Science and Artificial Intelligence Lab (CSAIL),  Massachusetts Institute of Technology, Cambridge, MA 02139 USA.
       }%
\thanks{$^{5}$Institute of Ocean Research, Peking University, Beijing, 100871, China.
        }%
\thanks{$^{6}$Peng Cheng Laboratory, Shenzhen 518055, China.
        }%
\thanks{$^{\ast}$Corresponding authors: 
       W. Wang ({\tt\small  wweiwang@mit.edu}); G. Xie ({\tt\small xiegming@pku.edu.cn})
        }
}
\begin{document}

\maketitle
\thispagestyle{empty}
\pagestyle{empty}

%%%%%%%%%%%%%%%%%%%%%%%%%%%%%%%%%%%%%%%%%%%%%%%%%%%%%%%%%%%%%%%%%%%%%%%%%%%%%%%%
\begin{abstract}
Underwater communication is extremely challenging for small underwater robots which typically have stringent power and size constraints. In our previous work, we developed an artificial electrocommunication system which could be an alternative for the communication of small underwater robots. This paper further presents a new electrocommunication system that utilizes Binary Frequency  Shift Keying (2FSK) modulation and deep-learning-based demodulation for underwater robots. We first derive an underwater electrocommunication model that covers both the near-field area and a large transition area outside of the near-field area. 2FSK modulation is adopted to improve the anti-interference ability of the electric signal. A deep learning algorithm is used to demodulate the electric signal by the receiver. Simulations and experiments show that with the same testing condition, the new communication system outperforms the previous system in both the communication distance and the data transmitting rate. In specific, the newly developed communication system achieves stable communication within the distance of 10 m at a data transfer rate of 5 Kbps with a power consumption of less than 0.1 W. 
The substantial increase in communication distance further improves the possibility of electrocommunication in underwater robotics.
%The proposed system provides a new theoretical basis for the long-distance underwater electrocommunication. At the same time, an effective solution is given in the design and implementation of the entire system, which further lays the foundation for the long-distance underwater electrocommunication.
\end{abstract}

\section{INTRODUCTION}

%Human soc requires the continuous supply of various resources. With the increasing tension of land resources, the ocean has become an important space for humans to obtain resources such as energy, minerals, and living things \cite{WangA}. Relying solely on human work can no longer keep up with the development needs of the times. Underwater robots have become new hotspots because they can replace humans in resource exploration, fishing, and mining of underwater dangerous conditions \cite{Yuh2000Design}.
%Many times, a single robot cannot meet the needs of the job, and complex tasks require the cooperation of multiple robots \cite{WergerCooperation}. 
In complex marine environments, efficient communication is very important. Underwater acoustic communication is one of the common communication methods, which however has problems of large Doppler shifts and multi-path effects \cite{KilfoyleThe, Yang2016Communication}. Underwater optical communication is easily affected by the water conditions and cannot be applied in turbid waters \cite{Hanson2008High, Che2011Re, Hasan2017Underwater}. Moreover, electromagnetic communication cannot be used underwater because electromagnetic waves are very seriously attenuated in water, and signals cannot be transmitted over long distances underwater \cite{Che2011Re}. Electrocommunication \cite{WeiICRA2015a}, as a new communication method inspired by weakly electric fish \cite{GabbianiFrom, Fleishman1992Communication}, can effectively avoid the above-mentioned shortcomings and attracts much attention.

Electrocommunication (also called electric field communication) has been successfully used in robots after decades of development. Electric field communication devices were first manufactured and tested in Japan \cite{Swain1970An}. The Japan Marine Science and Technology Research Center studied the principle of underwater electric field communication and derived a relatively accurate physical model \cite{Momma1976Underwater}. In China, Northwestern Polytechnical University designed a DSP-based electrocommunication device, which was an important practical achievement \cite{Wu2010A}. %Underwater robots are playing an increasingly important role in many large-scale operation tasks, such as resource detection, security inspection, search and rescue, and so on. 
Recently, an artificial electrocommunication system based on 2ASK modulation was developed for and successfully implemented in small underwater robots \cite{WangA, Yang2018Communication}. Thus, it is a
promising setup for communication within small underwater robots. However, the communication distance of the electrocommunication system in \cite{WangA, Zhang2017CSMA} is still very short for underwater robots, which limits its application in the practice environment. One possible solution to increase the communication distance is simply improving the transmitting power of the system. However, according to the physical model of electrocommunication \cite{Yang2016Communication}, the transmitting power grows the third power concerning the communication distance. Therefore, increasing the transmitting power could not be viable to dramatically enlarge the communication distance for miniature robots that have limited power and space.

To solve the problems of miniaturization, long-distance, and low power in electrocommunication, a new electrocommunication system is developed in this paper. We adopt 2FSK modulation to improve the anti-noise interference ability. Besides, by using a deep learning-based demodulation method the system can tolerate complex environmental noises, and improving the communication quality.

In particular, a new mathematical model is first proposed for electrocommunication. Compared with the dipole field model in \cite {WangA}, the distribution of the electric field outside the near field region and the influence of frequency on the receiver are further considered. Second, 2FSK modulation and deep learning-based demodulation are adopted in our system.  Comparisons with previous methods are performed in the simulation to demonstrate the improvement of the communication distance. Finally, the experiments verify the feasibility of transmitting underwater electric field signals at a distance of 10 m with a data transfer rate of 5 Kbps. The results show that the new communication system has the advantages of miniaturization, long-distance, and low power consumption compared with the previous version of the electrocommunication system in \cite{WangA}. 

\section{Principle of Electrocommunication}
\begin{figure}[thpb]
   \centering
   \includegraphics[scale=0.26]{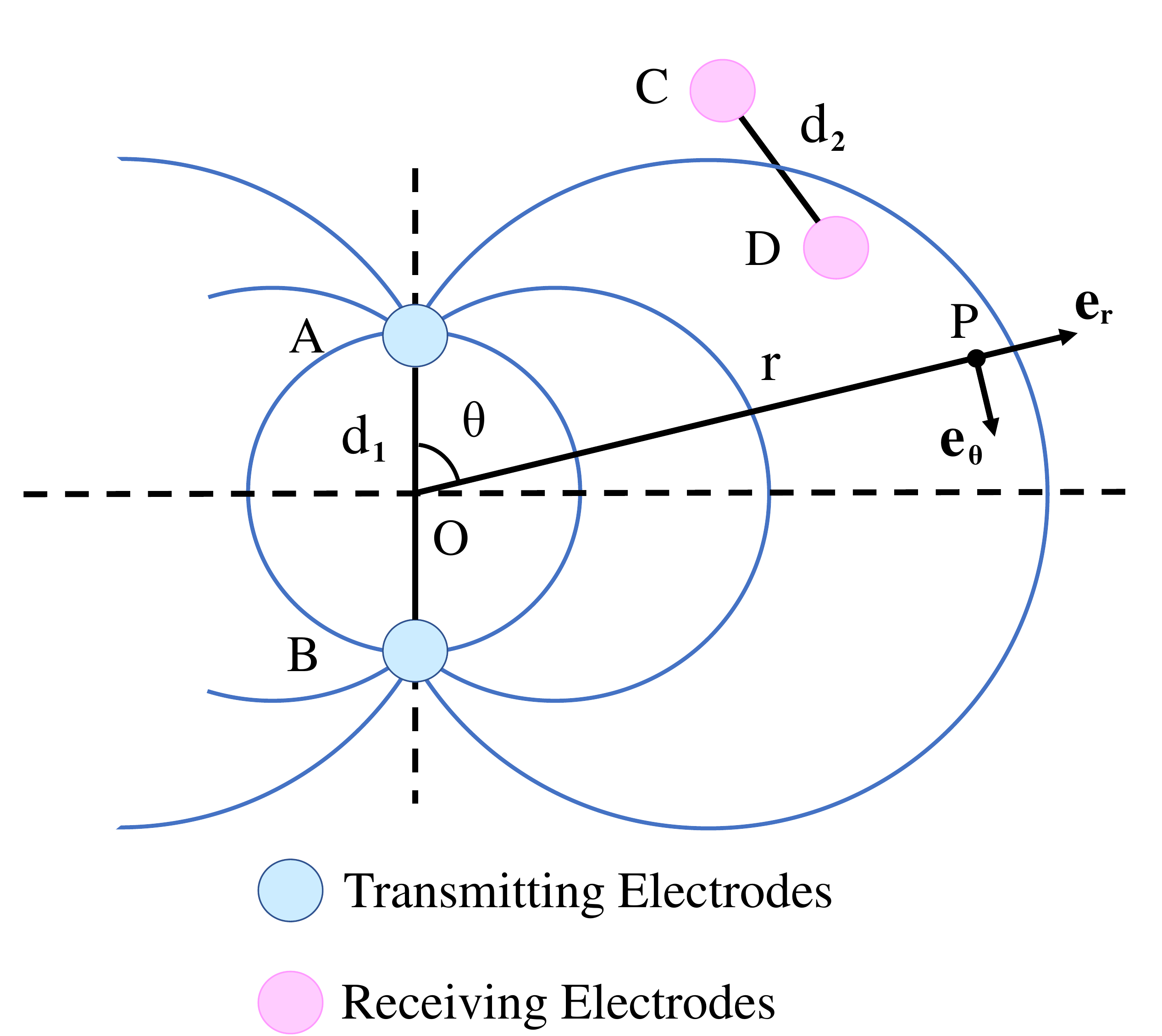}
   \caption{Overall concept of underwater electrocommunication. P is one point near the transmitting electrodes; $\textbf{e}_r$ and $\textbf{e}_\theta$ are unit vectors in the radial and azimuthal directions in polar coordinates; $d_1$ is the distance between the transmitting electrodes and $d_2$ is the distance between the receiving electrodes; r is the distance between point P and the midpoint of transmitting electrodes and $\theta$ is the polar angle of point P.}
   \label{overallconceptual}
\end{figure}

Underwater electrocommunication is a type of communication that uses the electric field as the information carrier and uses water as the transmission medium \cite{Momma1976Underwater}. As shown in Fig. \ref{overallconceptual}, underwater electric field communication is achieved through a pair of transmitting electrodes, and a pair of receiving electrodes.
The data is encoded, modulated, and further amplified by the transmitting electrode. The signal propagates through the current field in the conductive water and is then received by the receiving electrode. At the receiver side, the signal undergoes amplification and demodulation, and finally, the signal is decoded.

\subsection{Model of Electrocommunication}
During electromagnetic communications, the changing electromagnetic field will excite both the displacement current and the conduction current \cite{inproceedings}. When the frequency of the signal is lower than 230 MHz, the propagation of the signal in the water mainly depends on the conduction current, and the displacement current can be ignored. In this condition, we only consider the effect of the conduction current on the electric field, and we call it electrocommunication.

The electric field distribution in electrocommunication can be approximated as a dipole field model. Its principle is shown in Fig. \ref{overallconceptual}.
In particular, for a sinusoidal alternating electric field $\mathbf {E} (t) = \mathbf {E_0} \cos\omega t$ with an angular frequency of $ \omega $ under the water, the amplitude ratio of the displacement current $J_d$ and the conduction current $J_c$ can be expressed as
\begin{equation}
  \left|\frac{J_d}{J_c}\right| = \frac{\varepsilon}{\sigma}\omega
\end{equation}
where $ \varepsilon $ is the dielectric constant of water and $ \sigma $ is the electric conductivity of water. Studies show that at 20 $^{\circ} $C, the dielectric constant of water ($\varepsilon$ = 7.08$\times$10$^{-10}$ F$\cdot$m$^{-1}$) is 80 times that of air $ \varepsilon_0$ ($\varepsilon_0$ = 8.85$\times$10$^{-12}$ F$\cdot$m$^{-1}$). For seawater, $\sigma$ is about 4 S$\cdot$m$^{-1}$ while $\sigma$ is between  0.01 S$\cdot$m$^{-1}$ and 0.1 S$\cdot$m$^{-1}$ for freshwater. The displacement current can be neglected if (2) satisfies

\begin{equation}
\left|\frac{J_d}{J_c}\right| \ll 0.1
\end{equation}

The electrocommunication emitter at this time can be equivalent to a dipole model, as shown in {Fig. 1}. Using a cylindrical coordinate system, according to electrical theory, the electric field strength vector at the receiving end is defined as \cite{Wu2010}:
\begin{equation}
\mathbf{E_z}=\frac{\mu_0 I_0 d_1 e^{-jkr}}{4\pi r}\mathbf{e_z}
\end{equation}
Where $I_0$ stands for the voltage between transmitting electrodes. The electric field strength can be expressed as follows

\begin{equation}
\mathbf{E_r}(r,\theta)=\frac{I_0 d_1 \cos\theta}{2\pi \sigma} (\frac{1}{r^3}+j\frac{k}{r^2})e^{-jkr}\mathbf{e_r}
\end{equation}
\begin{equation}
\mathbf{E_\theta}(r,\theta)=\frac{I_0 d_1 \sin\theta}{4\pi \sigma} (\frac{1}{r^3}+j\frac{k}{r^2}-\frac{k^2}{r})e^{-jkr}\mathbf{e_\theta}
\end{equation}
\begin{equation}
k=\sqrt{\omega \mu \sigma/2}+j\sqrt{\omega \mu \sigma/2}
\end{equation}

In particular, when the axes of the transmitting electrode and the receiving electrode coincide, the field strength at the receiving electrode is
\begin{equation}
\mathbf{E}(r)=\frac{I_0 d_1}{4\pi \sigma}\sqrt{\frac{3d_2}{4r^2+{d_2}^2}+1} (\frac{1}{r^3}+j\frac{k}{r^2}-\frac{k^2}{r})\ e^{-jkr}
\end{equation}
The relationship between electric potential and field strength is
\begin{equation}
\varphi=\int \mathbf{E} \cdot \mathbf{dr}
\end{equation}
From (7)-(8), the potential at the receiving electrode can be obtained as follow
\begin{equation}
\varphi=\int_{0}^{d_2} \frac{I_0 d_1}{4\pi \sigma}\sqrt{\frac{3d_2}{4r^2+{d_2}^2}+1} (\frac{1}{r^3}+j\frac{k}{r^2}-\frac{k^2}{r})\ e^{-jkr}dr
\end{equation}
where $d_2$ stands for the distance between two receiving electrodes.
Furthermore, the voltage between the receiving electrodes C and D is
\begin{equation}
U_{CD}=\int_{0}^{d_2}E(r)\cdot dr=\int_{0}^{d_2}\frac{I_0 d_1}{4\pi \sigma} (\frac{1}{r^3}+j\frac{k}{r^2}-\frac{k^2}{r})\ e^{-jkr}dr
\end{equation}
In particular, when the distance between the transmitting end and the receiving end is relatively short ($ R \ll \frac{\lambda}{2 \pi} $), a region with a radius of R is generally defined as a near-field region. Where $ \lambda = 2\pi \sqrt{\frac{2}{\omega \mu \sigma}}  $ is the wavelength of the transmitted wave. When the frequency is high, the range of the near-field region is small. Therefore, if we want to increase the electric field communication distance, the theory of the near field is limited and cannot be used.

Let intermediate variable $ t = r\sqrt{\omega \mu \sigma/2} $ , substituting $t$ into (10) gives
\begin{equation}
U_2=\frac{I_0 d_1 d_2 }{4\pi \sigma r^3}\sqrt{1+2t+2t^2+4t^3+4t^4}\ e^{-t}
\end{equation}
$U_2$ represents the receiving voltage. It can be known from the above equation that the receiving voltage and the transmitting electrode plate, the distance between the receiving electrode plate, the current through the transmitting electrode, the conductivity of water, and the transmission distance of the electric field signal are all related. Generally, when the distance between the transmitting electrodes $ d_1 $ and the distance between the receiving electrodes is determined, the resistance of the water $ R_w $ is determined, and the equation can be transformed into:
\begin{equation}
U_2=\frac{U_1 d_1 d_2}{4\pi \sigma r^3 R_w}\sqrt{1+2t+2t^2+4t^3+4t^4}\ e^{-t}
\end{equation}

To further explore the influence of the receiving voltage with the communication distance and the carrier frequency outside the near field area.  Let the current between the transmitting electrodes $I_0$ = 0.5 A. The distance between the transmitting electrodes $d_1$ = 0.25 m while the distance between the receiving electrodes $d_2$ = 0.25 m. For freshwater, we let $\sigma$ =0.01 S$^{-1}$. When the distance $r$ between transmitting and receiving electrodes is over than 3m ($r > 10 \times d$, and $d$ is the larger of $ d_1 $ and $ d_2 $ ), it can be regarded as outside the near field area. The change of the receiving voltage is as {Fig. 2}.

\begin{figure}[thpb]
   \centering
\includegraphics[scale=0.4]{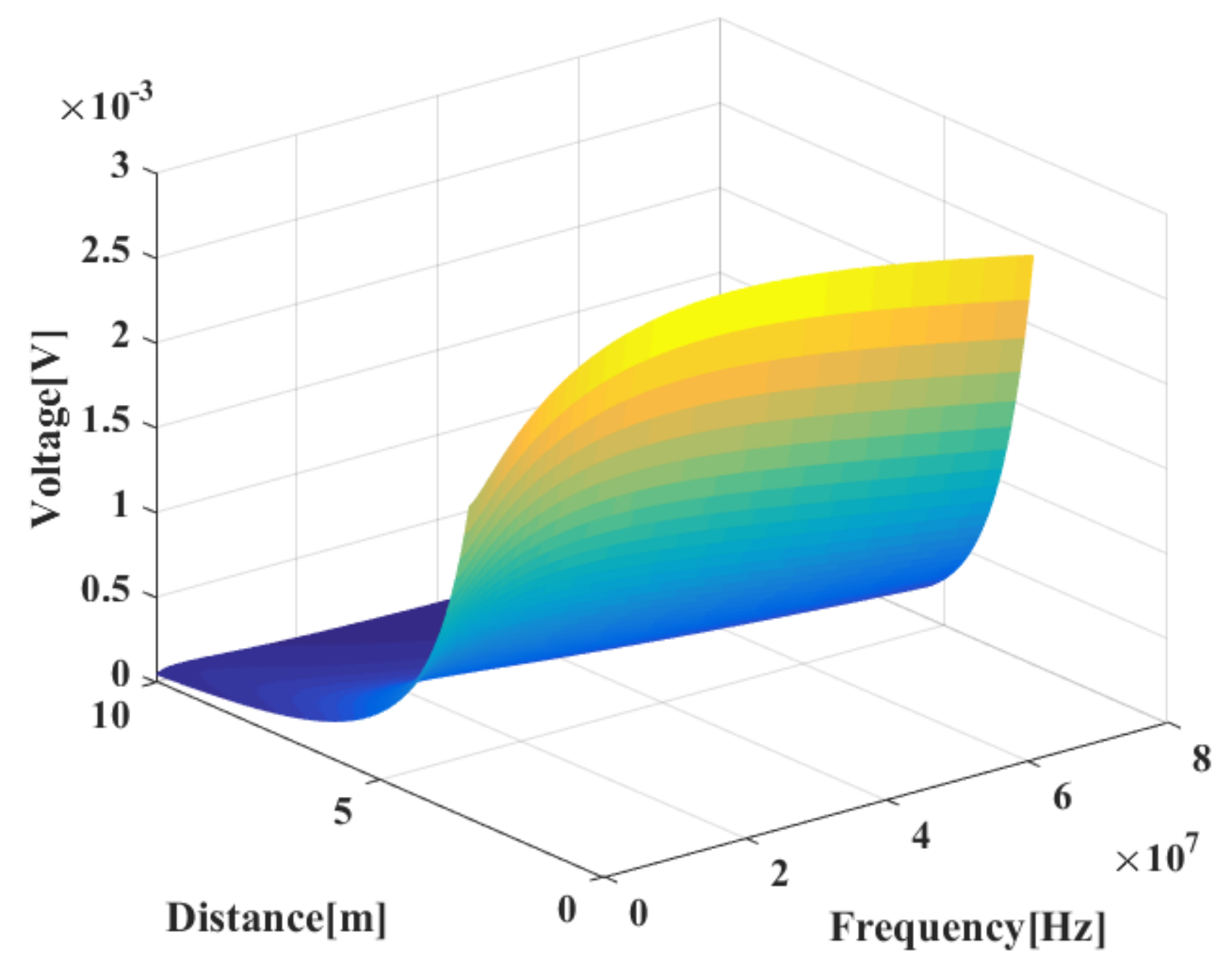}
   \caption{Relation curve of received voltage with distance and frequency.}
   \label{}
\end{figure}

\begin{figure}[thpb]
   \centering
\includegraphics[width=1\linewidth] {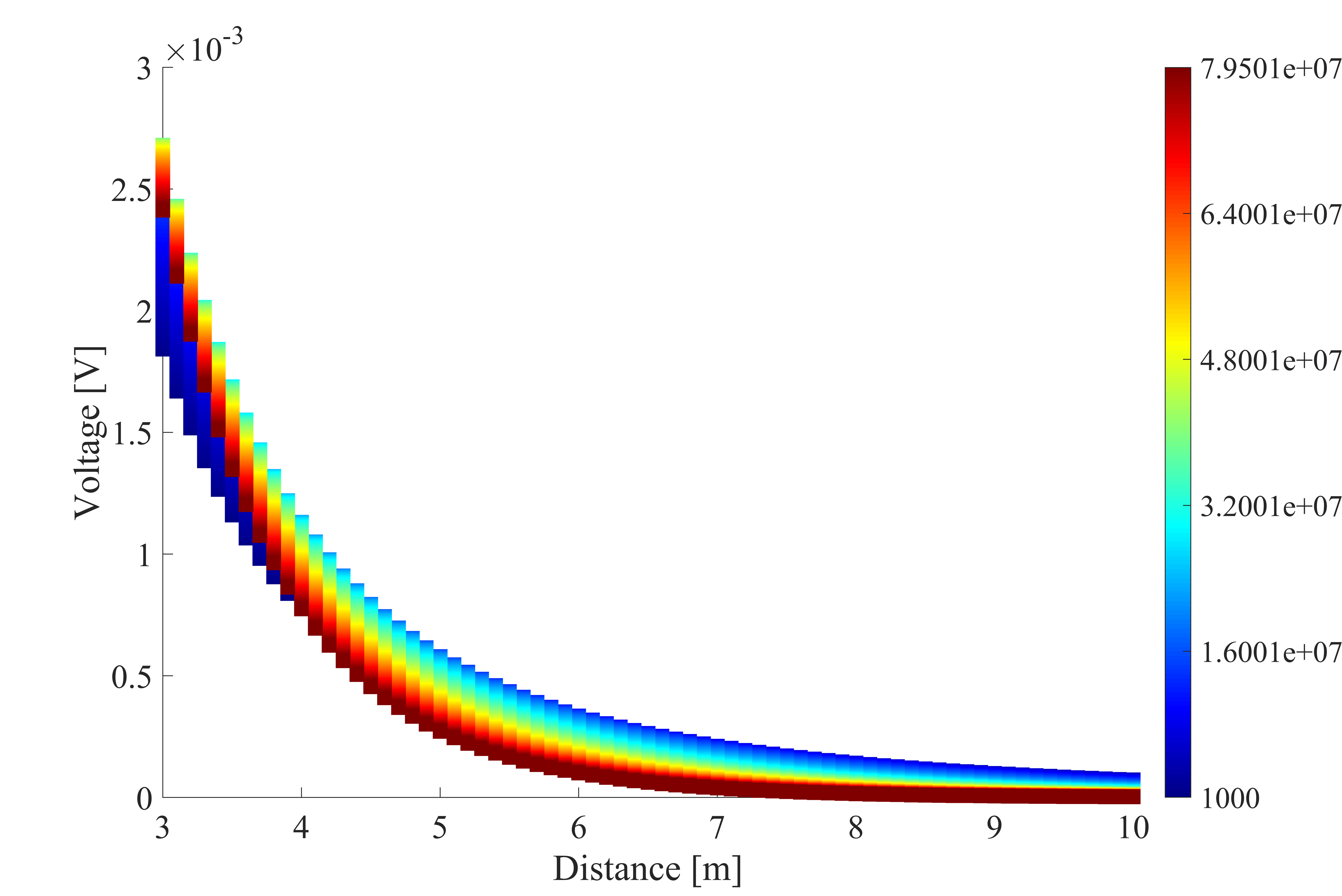}
   \caption{Relation curve of received voltage with distance. The unit of the colorbar on the right side is Hz.}
   \label{}
\end{figure}

In {Fig. 3}, the color bar represents the frequency distribution. It can be seen that with the increase of communication distance, the frequency factor has less and less influence on the receiving voltage. When the frequency is lower than 1 MHz, its effect on voltage is negligible. So it is feasible to conduct electrocommunication at a relatively high frequency. Moreover, we can see that the receiving voltage decreased greatly with the communication distance. Therefore, to increase the communication, we need to largely improve the system ability in detecting the weak voltage signals at the receiver side.

\section{NEW Electrocommunication System Design}

\subsection{Overall Concept Design}
%First, as the transmission signal of electrocommunication, voltage signal has a relatively strong transmission ability in the water. 

%Secondly, When designing this system, the software and hardware were fully considered. To make the system miniaturized and integrated, software filtering, modulation, and demodulation strategies are adopted to reduce the complexity of the hardware circuit and make it more concise and practical.

As shown in  {Fig. 4}, we use 2FSK as the signal modulation mode, which is easy to demodulate and helps reduce the bit error rate. In order to increase the communication distance, we also use deep learning methods during demodulation.
\begin{figure}[thpb]
   \centering
\includegraphics[scale=0.21]{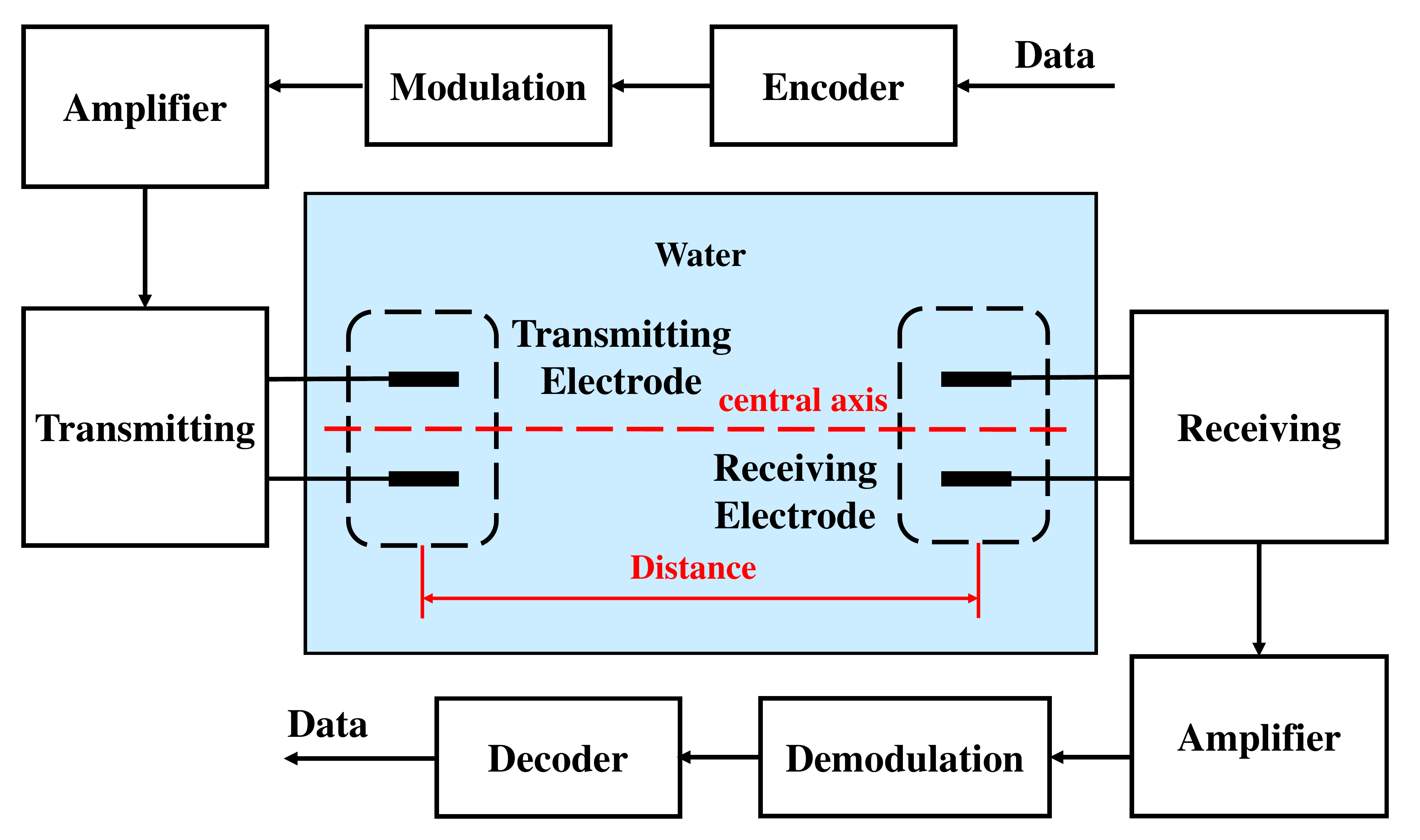}
   \caption{Overall design of the new electrocommunication system.}
   \label{}
\end{figure}

\subsection{Transmitting Unit Design}
The transmitting unit mainly includes power supply, microcontroller, digital-to-analog conversion module, amplifier, and transmitting electrode.
% The 7.4 V lithium battery is used to meet the power requirements. The main controller we use is Raspberry Pi 4B. For the digital-to-analog conversion, we use DAC902E. The amplifier is used to amplify the voltage signal and provide enough power to ensure that a stable voltage signal can be maintained between the electrode plates in the water.

There are many signal modulation methods, including binary amplitude shift keying (2ASK), binary frequency shift keying (2FSK), binary phase shift keying (2PSK), and so on \cite{Sharma2017VERILOG, Chong2012Digital}. Their working principles are illustrated in {Fig. 5}. 
\begin{figure}[thpb]
   \centering
\includegraphics[scale=0.16]{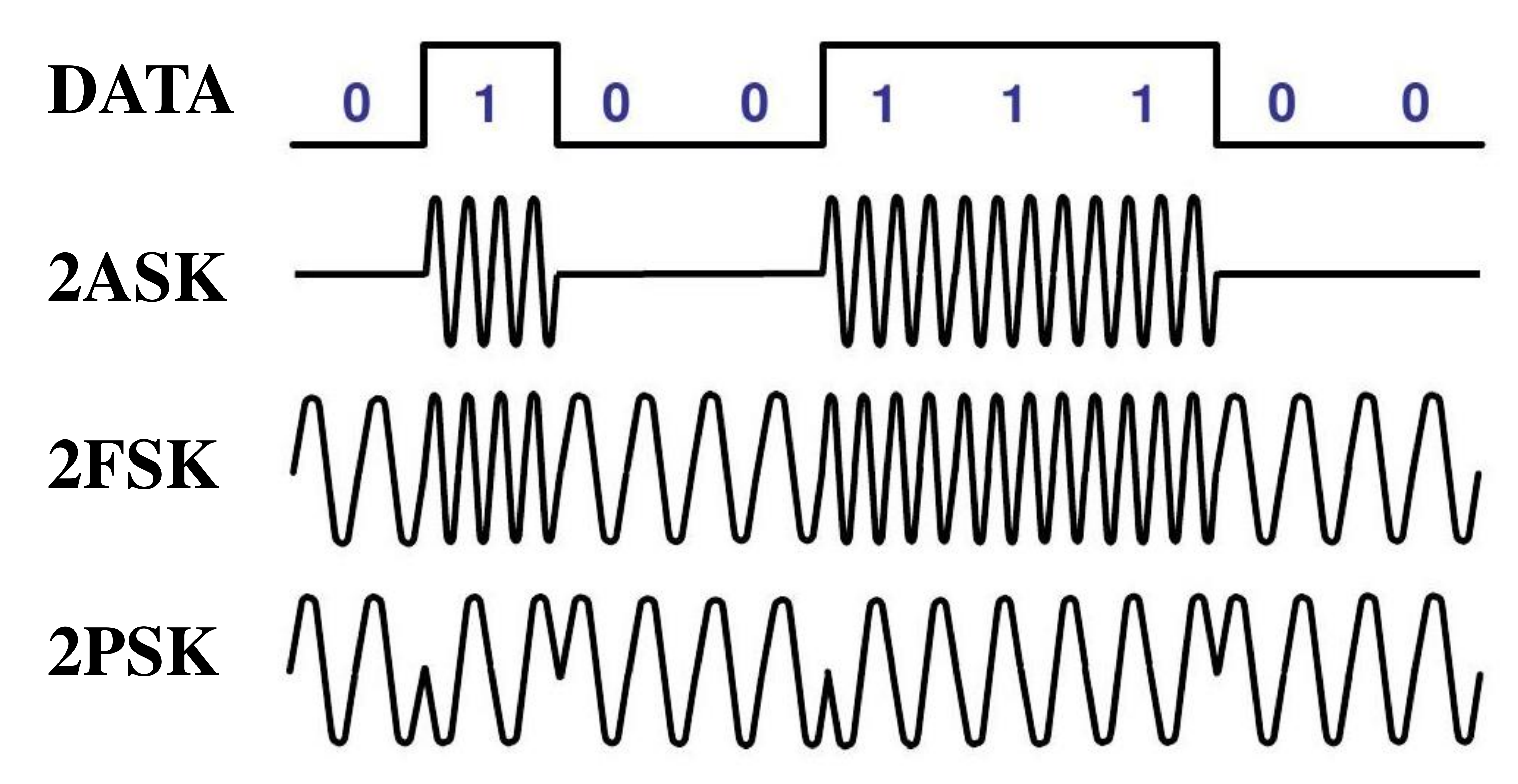}
   \caption{Several common modulation methods.}
   \label{}
\end{figure}
The advantage of ASK modulation is that the implementation is simple, but it is sensitive to the channel and has poor anti-interference ability. Although PSK modulation technology has strong anti-interference ability, it is complicated \cite{Marinovich1995Classification}. 2FSK combines the advantages of both anti-interference and low complexity, is adopted in our system.

2FSK is a binary digital frequency modulation that uses carrier frequency to transmit digital information \cite{Wang20172FSK}. In modulation, the transmitted digital information is used to control the frequency of the carrier wave, and two different frequency $ ( f_1$, $ f_2 ) $ waves are used to represent two different symbols(`0' and `1'). Sine waves are often used in actual communication systems. The change between $ f_1 $ and $ f_2 $ is instantaneous. Compared with the 2ASK modulation used in the electric field communication in \cite{WangA}, 2FSK is not sensitive to channel changes, so it is more resistant to interference.

The transmission unit is implemented and shown in {Fig. 6}. 
The 2FSK modulation method uses multiple carriers to carry one symbol information, which is beneficial to reduce the bit error rate after fading through a long distance. First, the data is converted to binary, and then we use 2FSK modulation to convert the data into an electric field signal so it can be transmitted in the underwater channel. As a result, the data is converted into a voltage signal for transmission and received from a distance.
\begin{figure}[htpb]
   \centering
\includegraphics[scale=0.25]{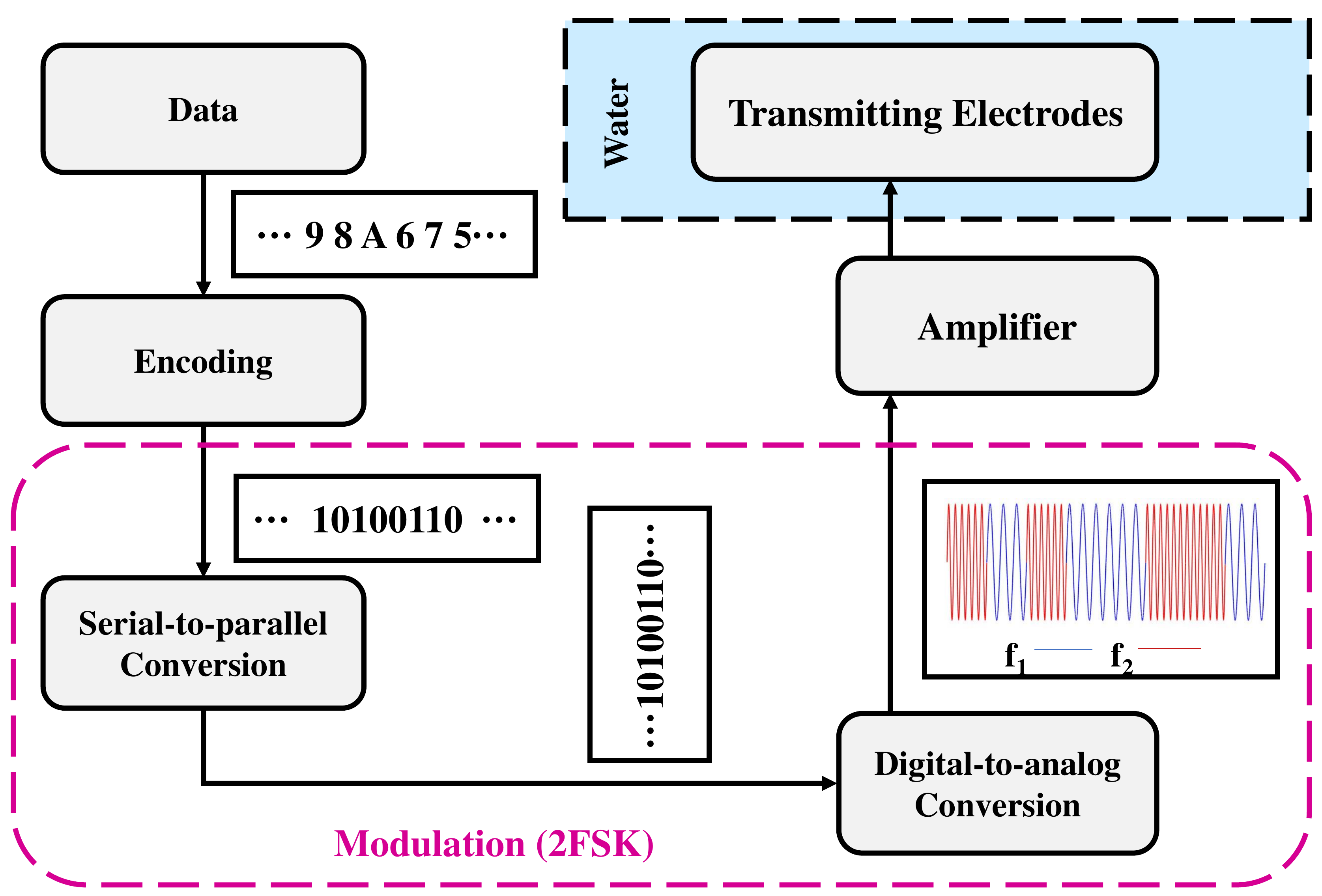}
   \caption{The transmitting unit.}
   \label{}
\end{figure}

 After data encoding, serial-to-parallel conversion, analog-to-digital conversion and 2FSK modulation, the waveform of the transmitted signal is obtained. The transmitting signal passes through the amplifier and is transmitted into the water by a pair of transmitting electrodes.

\subsection{Receiving Unit Design}
The receiving unit mainly includes the receiving electrode, amplifier, digital-to-analog conversion module, and main controller. The receiving electrode has the same structure as the transmitting electrode. The square electrode plate has a larger area, which is conducive to acquire signals. After a long distance, the electric field signal is severely attenuated, and the amplifier amplifies the signal to an intensity level sufficient to be recognized by the detection device. The amplified signal passes through the analog-to-digital converter and then processed by software filtering and demodulation to recover the data.
%In order to implement the digital-to-analog conversion, the AD9220 is adopted. The input of the AD9220 is highly flexible, allowing for easy interfacing to communications and data-acquisition systems. A truly differential input structure allows for both single-ended and differential input interfaces of varying input spans, with excellent performance.

For demodulation, a kind of deep learning method is used. The demodulation of the 2FSK modulation signal is regarded as a binary classification problem. 
%In actual operation, a sequence of sampling points is corresponding to symbol `0' or `1'. The neural network has the advantages of strong generalization ability and a small amount of calculation, and it is suitable for underwater electric field communication under strong electric field interference. 
The schematic of neural network for demodulation is shown in {Fig. 7}.
\begin{figure}[thpb]
   \centering
\includegraphics[scale=0.25]{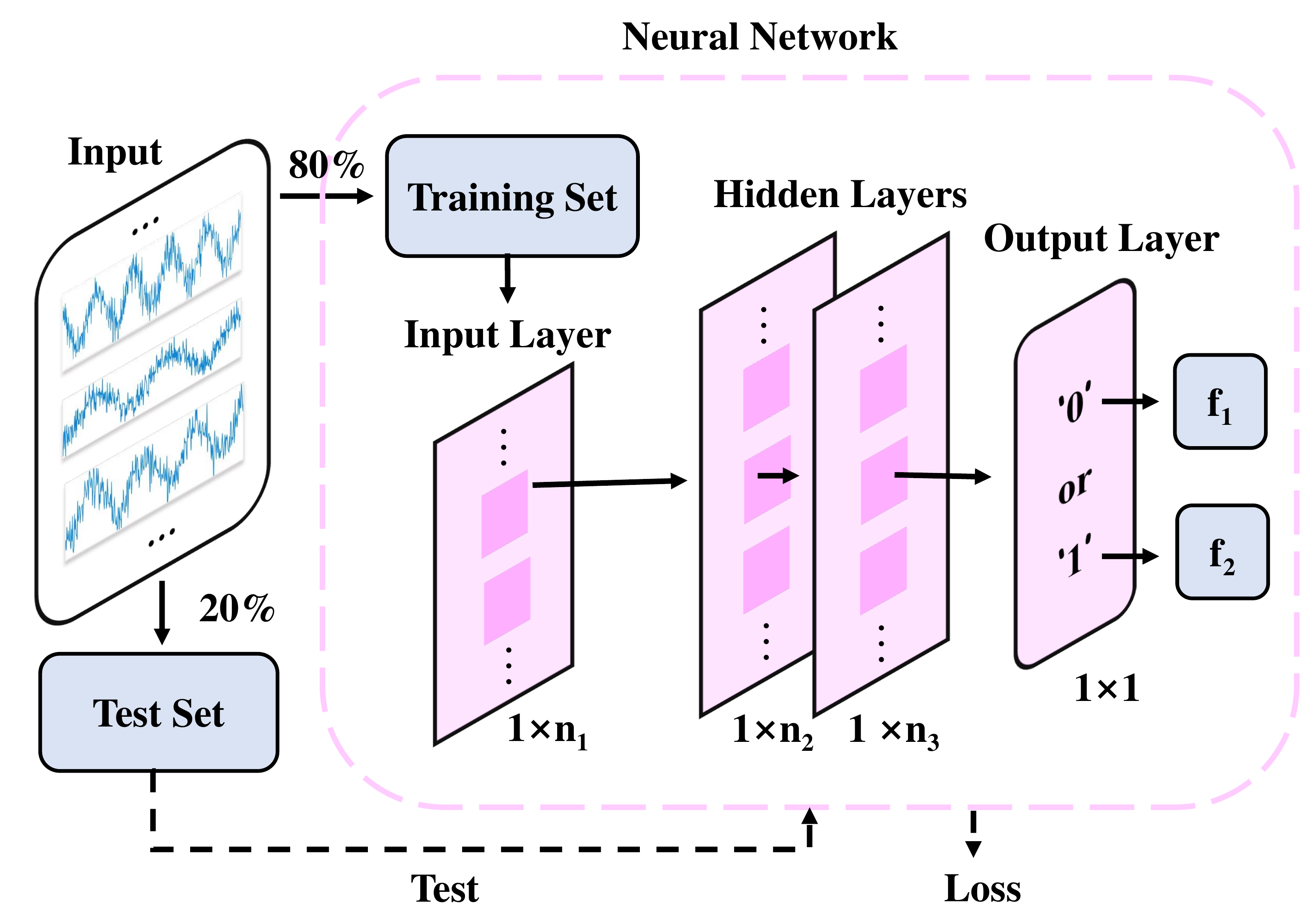}
   \caption{Schematic of neural network for demodulation.}
   \label{}
\end{figure}
%The rapid development of deep learning has promoted the improvement of computer and communication industries. Deep learning method is widely used in signal recognition, face detection, target acquisition and other fields, with high precision \cite{Sannino2018A}.

In particular, the neural network used for demodulation consists of an input layer, 2 hidden layers, and an output layer. The neural network has four layers in total, and each layer contains 30, 28, 10, and 1 neurons respectively. Increasing the number of neurons in the hidden layer can increase the network's spatial expression ability, which obtains better classification results. Inputting modulated signals with different distributions and different signal-to-noise ratios, and training the network with signals collected at different sampling rates can improve the generalization ability of the network. In practice, it is significantly more accurate than traditional coherent demodulation methods. The bit error rate is defined as the proportion of the bit errors in all transmitted bits to measure communication quality.

\begin{figure}[htp]
   \centering
\includegraphics[scale=0.63]{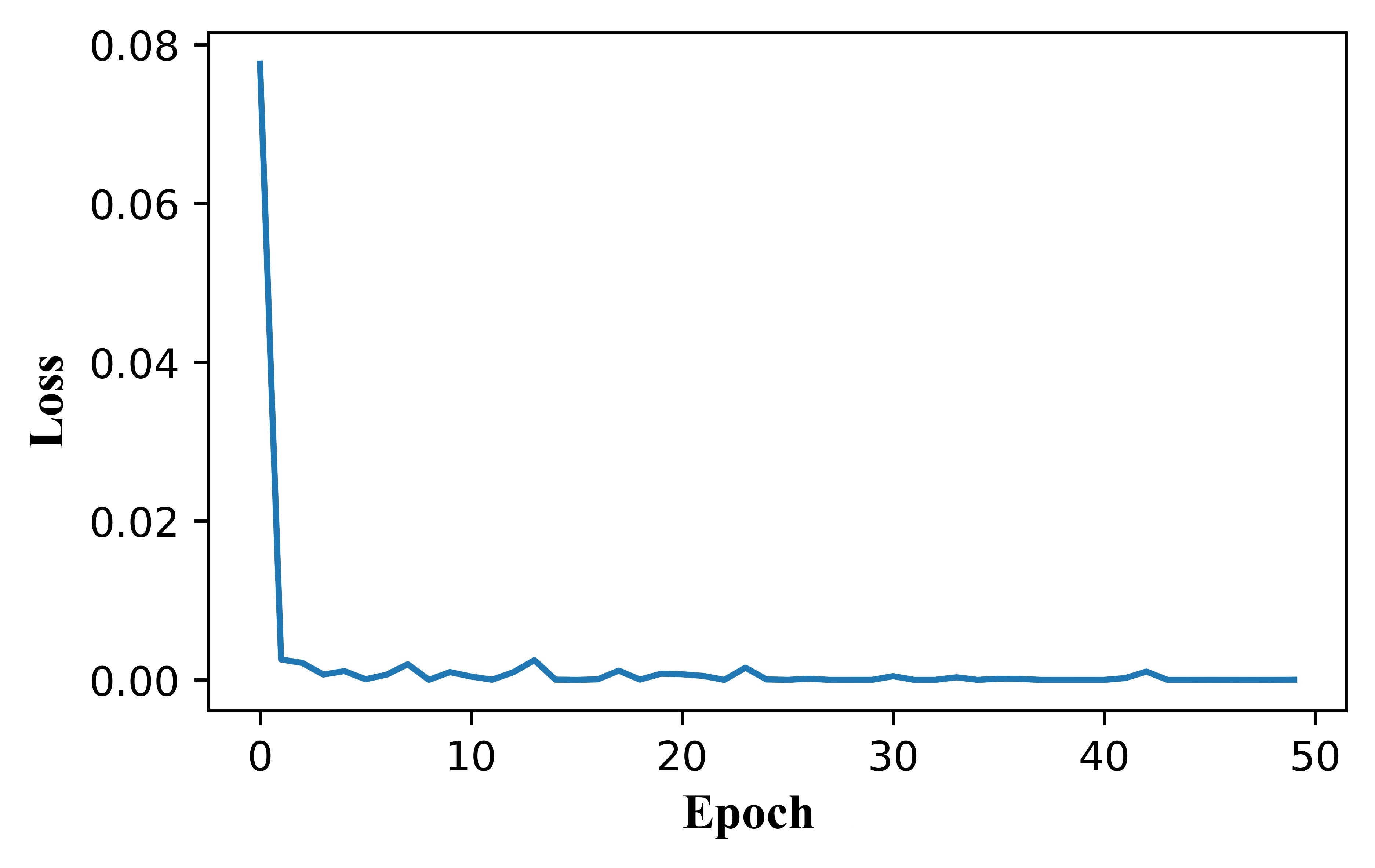}
   \caption{Loss in training neural networks (SNR = 0).}
   \label{}
\end{figure} 

\begin{figure}[htp]
   \centering
\includegraphics[scale=0.65]{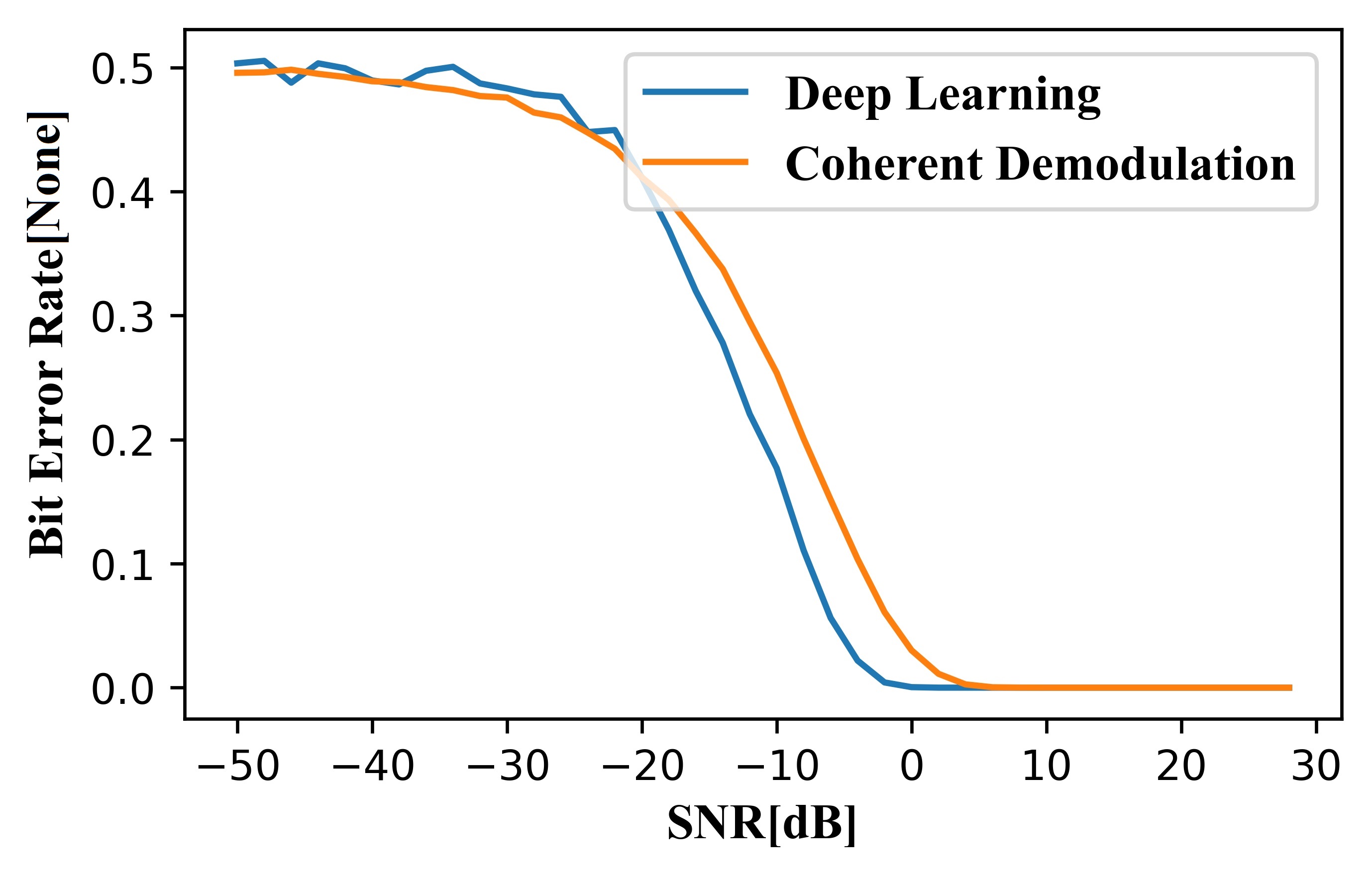}
   \caption{Comparison of different demodulation methods of 2FSK.}
   \label{}
\end{figure}

The reason why the neural network can play a better demodulation role is that it can converge quickly. As shown in {Fig. 8} above. The simulation results in the {Fig. 8} and {Fig. 9} show that when the signal-to-noise ratio $ (SNR) $ is greater than 2 dB, the bit error rate of demodulation using the deep learning method is zero. When the signal-to-noise ratio is less than about -5 dB, the bit error rate is very high and changes rapidly. For the signal-to-noise ratio in the range of -20 dB to 0 dB, deep learning is significantly better than traditional coherent demodulation. Even within a certain range, the bit error rate can be reduced by 10\%. As the signal-to-noise ratio decreases, the bit error rate greatly increases. This method is suitable for underwater electronic communication under most conditions. On the other hand, as the signal-to-noise ratio increases, deep learning demodulation can quickly reduce the bit error rate to zero (about SNR = -2 dB), while the traditional coherent demodulation method requires a signal-to-noise ratio greater than 5 dB to ensure the bit error rate is 0.

\begin{figure}[htp]
   \centering
\includegraphics[scale=0.65]{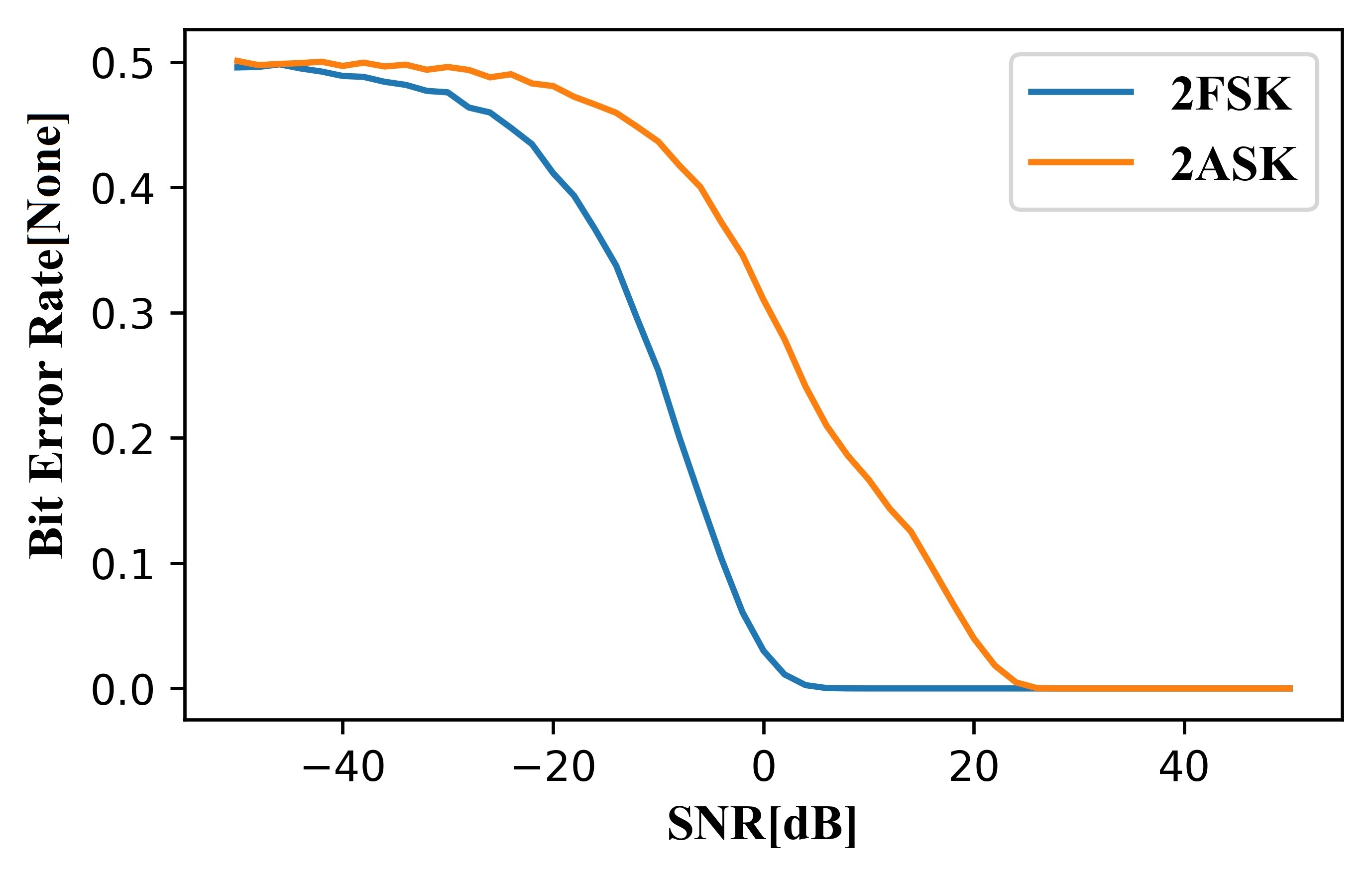}
   \caption{Influence of different modulation methods on communication bit error rate.}
   \label{}
\end{figure}

Compared with 2ASK modulation, 2FSK modulation has advantages in reducing bit error rate. The simulation results are shown in {Fig. 10}. It fully demonstrates the feasibility of 2FSK modulation in reducing the noise level and improving the distance of electrocommunication.
\section{Underwater Electrocommunication Experiment}
The experiments were performed in a 50-meter long and 25-meter wide pool. 
% To make the experiment more intuitive, a pair of communication electrode plates are used for transmission and reception.
The main purpose of this experiment is to verify that the new electric field communication system designed by us can increase the data transfer rate and the communication distance. During the experiment, the transmitting electrode and the receiving electrode were always on the same axis, and 
data was continuously transmitted and received at different distances. The transmitting device is placed on the shore, and only the transmitting electrode and the receiving electrode are immersed in water to a depth of 0.5 meters. Depth has little effect on communication, so we chose a convenient location. Since the receiving part is placed on the boat, we use the power bank to supply power and measure the communication distance with a long tape measure.  Every five meters a test point is set during the experiment, and the bit error rate is calculated. The experimental site and equipment are shown in {Fig. 11}.

\begin{figure}[htp]
   \centering
\includegraphics[scale=0.3]{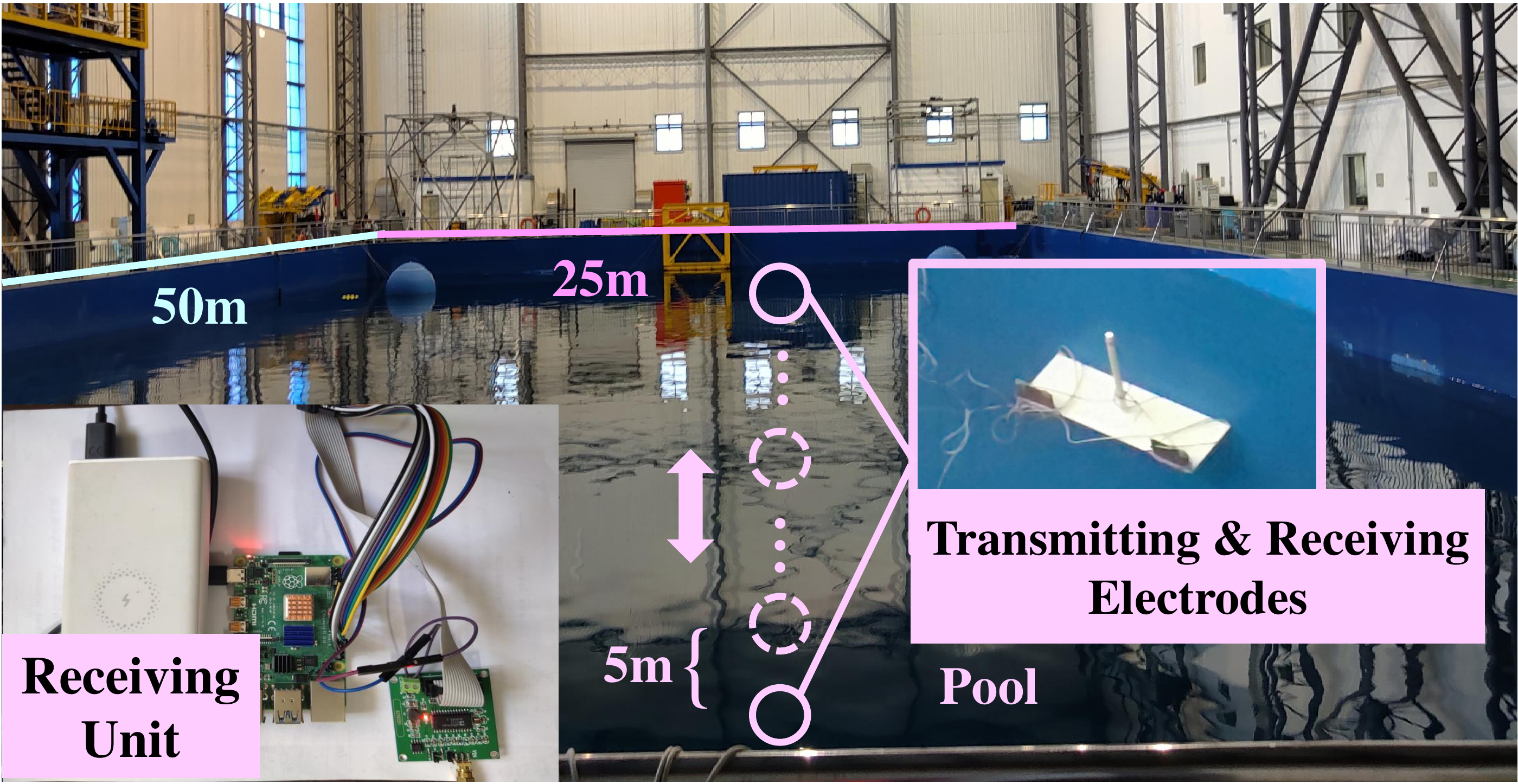}
   \caption{The test experiment of models. }
   \label{}
\end{figure}

We first verify the relationship between the experimental results and the theoretical formula we derived. In particular, we use a constant voltage source at the emitter side and test the value of the received voltage at different distances. Then, according to the theoretical model we derived, the unknown parameters in the model are solved. Finally, according to the model we solved, the theoretical value of the received voltage at each test point in the field is predicted.
% To obtain a more accurate model, the ratio of the transmitting voltage to the receiving voltage is measured in experiments as a key part of the environmental characteristics.
The transmitting terminal is connected with a constant voltage source, and the transmitting voltage is 40 V. In the voltage test, we take a test point every 0.2 m, and measure the received voltage within 1.6 m.
 As shown in Fig. 12, the electrocommunication model we established can well describes the distribution of the receiver's voltage in the environment. 
 \begin{figure}[htp]
   \centering
\includegraphics[scale=0.35]{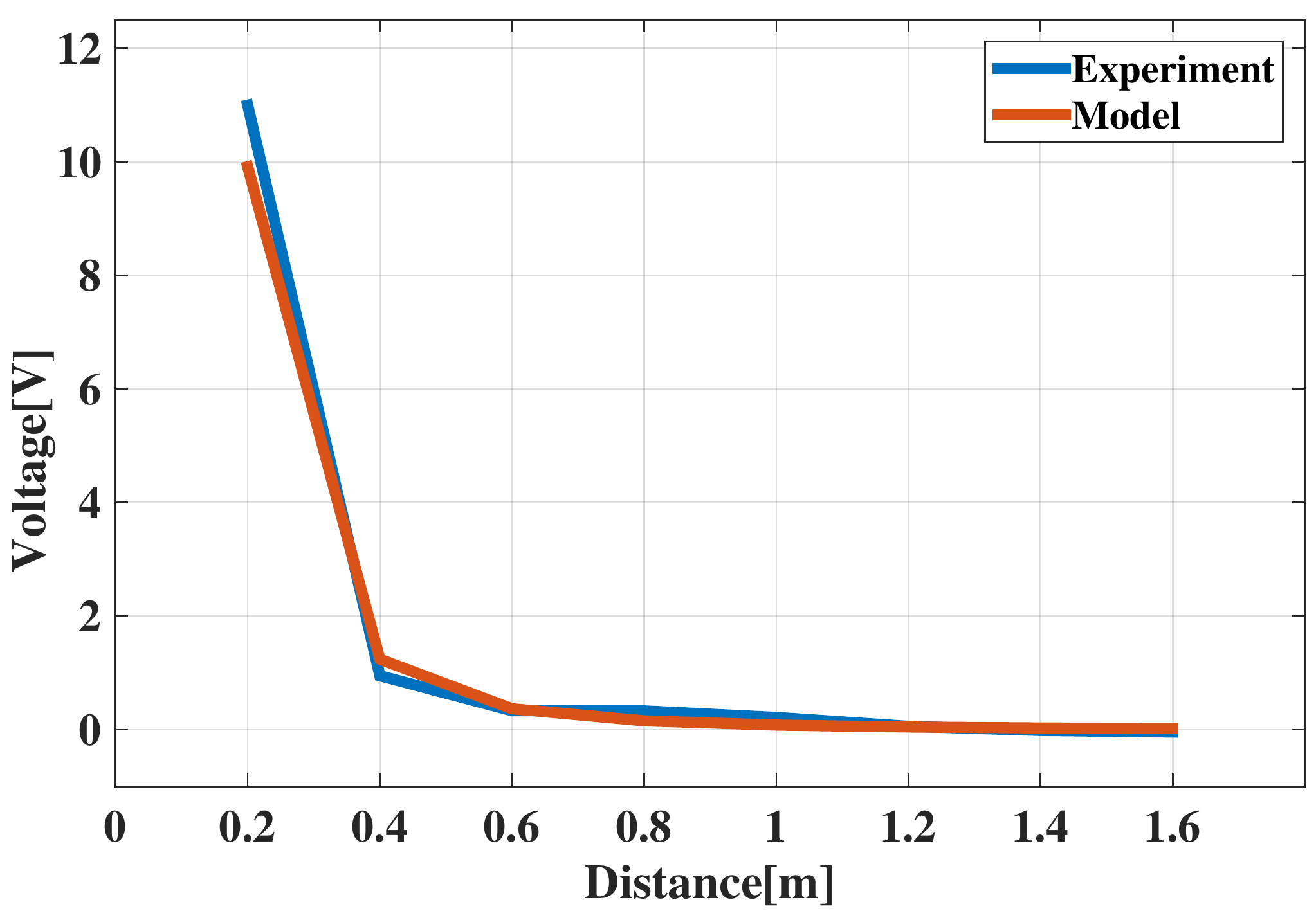}
   \caption{The receiver's voltage with distance using the designed electrocommunication system.}
   \label{}
\end{figure}

 In the communication experiment, we take a test point every 5 m. According to the received signal of each point, the bit error rate can be calculated. The actual emission voltage is set to be   V$_{pp}$ = 12V in the experiment. The transmitting power is described as following 
 
\begin{equation}
P_{t}=\frac{1}{T}\int_{0}^{T}(\frac{V_{pp}}{2}\sin(2\pi ft))^2 dt \cdot \frac{1}{R_{w}}=\frac{V_{pp}^2}{8R_{w}}<0.1 W
\end{equation}

As shown in Fig. 13, the bit error rate is zero when the communication distance is less than 10 meters.
\begin{figure}[htp]
   \centering
\includegraphics[scale=0.35]{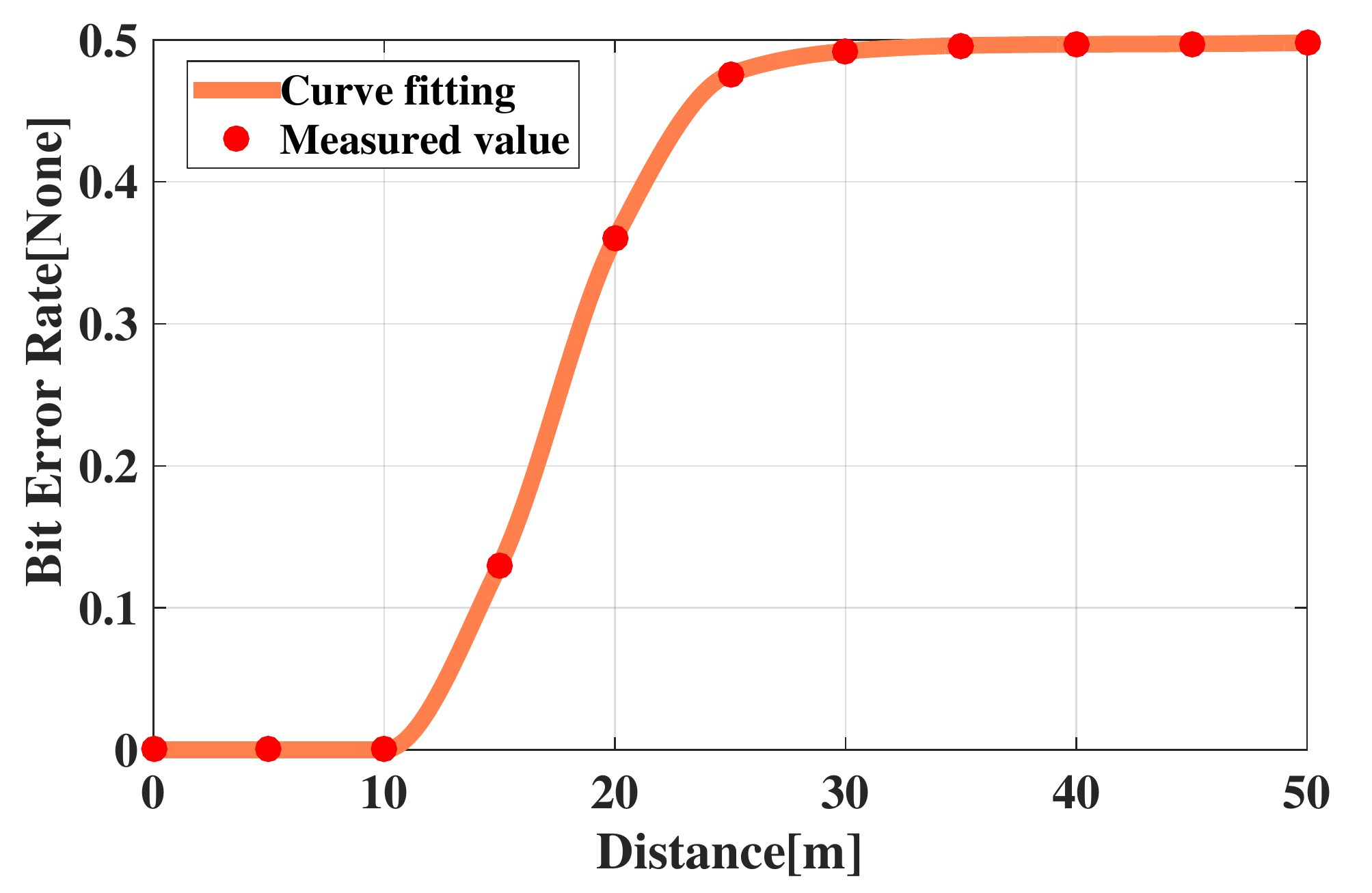}
   \caption{Bit error rate with distance using the designed electrocommunication system in the experiments.}
   \label{}
\end{figure}
When the distance is larger than 10 meters,  the bit error rate increases rapidly as the distance increases. By contrast, the previous demodulation system in \cite{WangA} cannot work because the number of  `0'  and `1' in transmitted data is similar.
In summary, the experiments demonstrate that the new electrocommunication system can communicate at a data transfer rate of 5 Kbps within 10 m,  implying that the new modulation and demodulation mechanism are good options for electrocommunication.

\section{CONCLUSIONS}

In this paper, a new electrocommunication system is designed based on 2FSK modulation a deep learning-based demodulation. A series of simulations and experiments have shown that the communication distance increases to more than 10 meters.
The electrocommunication system based on 2FSK modulation shows its advantages such as short time delay, low bit error rate, and low power in long-distance transmission. The development of underwater robots equipped with this system can improve battery life. It is also useful for future large-scale underwater cooperative communications of robots. However, the frequency band utilization rate of 2FSK modulation is still limited, and there is a lot of room for improvement. The use of frequency division multiplexing is a feasible direction in the future. Also, research on optimizing the deep learning framework to reduce the bit error rate is also meaningful.

%\addtolength{\textheight}{-12cm}   % This command serves to balance the column lengths
                                  % on the last page of the document manually. It shortens
                                  % the textheight of the last page by a suitable amount.
                                  % This command does not take effect until the next page
                                  % so it should come on the page before the last. Make
                                  % sure that you do not shorten the textheight too much.
%%%%%%%%%%%%%%%%%%%%%%%%%%%%%%%%%%%%%%%%%%%%%%%%%%%%%%%%%%%%%%%%%
\section*{ACKNOWLEDGMENT}

This work was supported in part by grants from the National Natural Science Foundation of China (NSFC, No.91648120, 61503008, 61633002, 51575005), the China Postdoctoral Science Foundation, (No. 2015M570013, 2016T90016), and the National Key R$\&$D Program of China (No. 2017YFB1400800).

\bibliographystyle{IEEEtran}
\bibliography{ElectrocommunicationIROS2020FINAL}
\end{document}